# Texture Classification Approach Based on Combination of Edge & Co-occurrence and Local Binary Pattern

**Shervan Fekri Ershad**
Department of computer science, engineering and IT, University of Shiraz, Shiraz, Iran
Shfekri@shirazu.ac.ir

**Abstract** – *Texture classification is one of the problems which has been paid much attention on by computer scientists since late 90s. If texture classification is done correctly and accurately, it can be used in many cases such as Pattern recognition, object tracking, and shape recognition. So far, there have been so many methods offered to solve this problem. Near all these methods have tried to extract and define features to separate different labels of textures really well. This article has offered an approach which has an overall process on the images of textures based on Local binary pattern and Gray Level Co-occurrence matrix and then by edge detection, and finally, extracting the statistical features from the images would classify them. Although, this approach is a general one and is could be used in different applications, the method has been tested on the stone texture and the results have been compared with some of the previous approaches to prove the quality of proposed approach.*

## 1 Introduction

Since 1990s, and with the progress of the computer vision and image processing, texture classification turned out to be one of the main subjects in the literature of these sciences. As texture classification has a close relationship with the sciences such as machine learning, it functions in areas like pattern recognition, object tracking, defect detection, face tracking and etc. The main problem in texture classification relates to two topics: 1.finding and introducing the best features for texture description 2.selecting the best and the most corresponding kinds of classifiers with selected features. Regarding these two topics, we have witnessed different kinds of discussions. Some of these methods directly work on the taken images of the textures, such as Texture Classification Based on Random Threshold Vector Technique [1], and Texture Classification Based on Primitive Pattern Units [3]. Another group of methods, first do some process on the images and then search for suitable features related to the class labels, such as texture classification by using advanced local binary patterns, and spatial distribution of dominant patterns [10] and A New Color-Texture Approach for Industrial Products Inspection [4]. Since the methods which have been offered up to now are applicative, and are appropriate per cases, therefore, they are not guaranteed to work effectively on other applications. For this reason, the researcher's main purpose is to define the new methods and features.

In this article, too, there has been a new method offered which is functioning as a general one. It can use for different applications, but works well on stone texture classification problem and can classify kinds of stone textures accurately by using a train step. In this method, first, the image is processed using LBP and GLCM algorithms. Then the edge detection has been done on the image by using edge filters. We can make a suitable data set by extracting famous statistical features like energy, entropy and contrast. Finally, we can classify the test data highly accurately by using classifiers.

In the result part, by using this approach, we have made suitable dataset for three different stone models called Granite, Travertine, and hatchet stone. Then by using some classifiers like NaiveBayes, KNN, and LADTree, we compute the misclassification error rate of this dataset. Also, we have compared it with some of the previous methods like LBP and GLCM regarding their accuracy rate. One of the main advantages of this approach is its high correspondence with the most classifiers which have been clearly stated in the results part.

### 1.1 Paper Organization

The reminder of this paper is organized as follows: Section two is related to the description of Local binary pattern (LBP) algorithm and the way of this estimation. Section three is related to the description of Gray level Co-occurrence matrix (GLCM) algorithm. Section four has a brief description about the different aspects of the edge detection filters. Section five describes the main method of this paper and finally, the results and conclusion included.

## 2 Local Binary Pattern

One of the most popular texture analysis operators is Local Binary pattern (LBP) that was introduced in [7]. It is a gray-scale invariant texture measure computed from the analysis of a 3x3 local neighborhood over a central pixel. The LBP is based on a binary code describing the local texture pattern. This code is built by thresholding a local neighborhood by the gray value of its center.

The eight neighbors are labeled using a binary code {0, 1} obtained by comparing their values to the central pixel value. If the tested gray value is below the gray value of the central pixel, then it is labeled 0, otherwise it is assigned the value 1:

$$d_i' = \begin{cases} 0 & if\ I(x_i, y_i) < I(x_0, y_0) \\ 1 & Otherwise \end{cases} \quad (1)$$

$d_i'$ is the obtained binary code, $d_i$ is the original pixel value at position $i$ and $d_0$ is the central pixel value. With this technique there is 256 ($2^8$) possible patterns (or texture units).

The obtained value is then multiplied by weights given to the corresponding pixels. The weight is given by the value $2_{i-1}$. Summing the obtained values gives the measure of the LBP:

$$I_{LBP} = \sum_{i=1}^{8} d_i' 2^{i-1} \quad (2)$$

Fig 1 shows an example on how to compute LBP. The original 3x3 neighborhood is given in Fig 1 (a). The central pixel value is used as a threshold in order to assign a binary value to its neighbors. Fig 1 (b) shows the result of thresholding the 3x3 neighborhood.

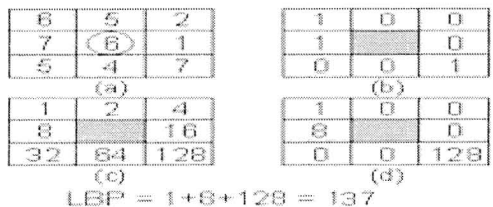

Figure1. Computation LBP

The obtained values are multiplied by their corresponding weights. The weights kernel is given by Fig 1 (c). The result is given in Fig 1 (d). The sum of the resulting values gives the LBP measure (137). The central pixel is replaced by the obtained value. A new LBP image is constructed by processing each pixel and its 3x3 neighbors in the original image. Fig 2 shows an example of the resulting LBP images.

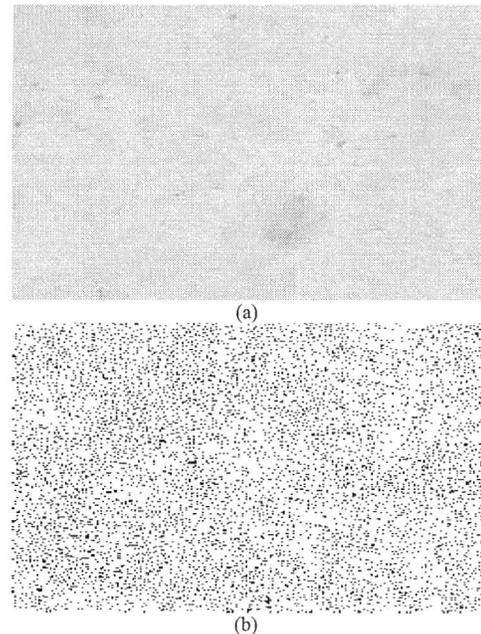

Figure2. Processing input image (a) Original Image of Travertine stone (b) LBP image

## 3 Gray Level Co-occurrence Matrix

Gray Level Co-occurrence Matrix (GLCM) was proposed in [6] by Haralick and Shanmugam. It is very useful in texture analysis. It calculates the second order statistics related to image properties by considering the spatial relationship of pixels. GLCM depicts how often different combinations of gray levels co-occur in an image. The GLCM is created by calculating how often a pixel with the intensity value i occurs in a specific spatial relationship to a pixel with the value j. The spatial Relationship can be specified in different ways, the default one is between a pixel and its immediate neighbor to its right. However we can specify this relationship with different offsets and angles. The pixel at position (i,j) in GLCM is the sum of the number of times the (i ,j) relationship occurs in the image.

Fig 3 describes how to compute the GLCM. It shows an image and its corresponding co-occurrence matrix using the default pixels spatial relationship (offset = +1 in x direction). For the pair (2,1) (pixel 2 followed at its right by pixel 1), it is found 2 times in the image, then the GLCM image will have 2 as a value in the position corresponding to $I_i$ =1 and $I_j$ =2. The GLCM matrix is a 256x256 matrix; $I_i$ and $I_j$ are the intensity values for an 8bit image. The GLCM can be computed for the eight directions around the pixel of interest (Fig 4). Summing results from different directions lead to the isotropic GLCM and help achieve a rotation invariant GLCM (Fig 5). More details are in [4].

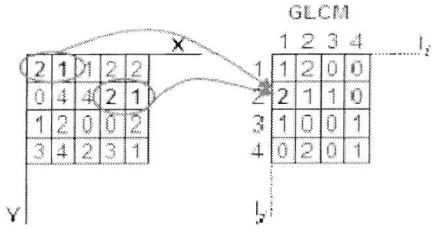

Figure3. Description of the Gray Level Co-occurrence Matrix.

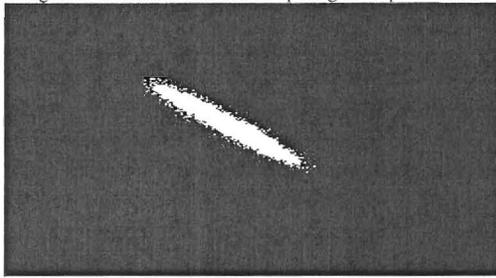

Figure4. Directions used for computing isotropic GLCM.

Figure5. Processing GLCM of Original Image of Travertine stone in 0'

## 4 Edge Detection

Edge detection is the process of localizing pixel intensity transitions. The edge detection has been used by object recognition, target tracking, segmentation, and etc. Therefore, the edge detection is one of the most important parts of image processing. There mainly exist several edge detection methods (Sobel [11], Prewitt [12], Roberts [13]). These methods have been proposed for detecting transitions in images. Early methods determined the best gradient operator to detect sharp intensity variations [14]. Commonly used method for detecting edges is to apply derivative operators on images. Derivative based approaches can be categorized into two groups, namely first and second order derivative methods. First order derivative based techniques depend on computing the gradient several directions and combining the result of each gradient. The value of the gradient magnitude and orientation is estimated using two differentiation masks [15].

In this work, Sobel which is an edge detection method is considered. This method is preferred will use in this work. The Sobel edge detector uses two masks, one vertical and one horizontal. These masks are generally used 3×3 matrices. Especially, the matrices which have 3×3 dimensions are used in matlab (see, edge.m).

## 5 Proposed method

The approach, which has been proposed in this article, is in a main box consisting of three sub-boxes shown in Fig 6. In this approach, first of all, according to the algorithm described in section 2, the LBP image would be estimated for the input image. The LBP's image is then would be sent to the second sub-box. In this sub-box, according to the algorithm mentioned in section 3, the second order statistics which is called GLCM, would be estimated for the second sub-box's input. GLCM can be estimated in 8 directions and achieve one image in every direction. Afterward, the GLCM images would be sent into the third sub-box. In the last sub-box using Sobel filter, the edges of entering images would be recognized. The edge detected images then would be sent out from the main box. In this point, every image which is out of the main box has statistical features by using the following formulas:

Entropy: $f_1 = \sum_{i,j=0}^{N-1} -Ln(P_{i,j})P_{i,j}$ (3)

Energy: $f_2 = \sum_{i,j=0}^{N-1}(P_{i,j})^2$ (4)

Contrast: $f_3 = \sum_{i,j=0}^{N-1} P_{ij}(i-j)^2$ (5)

Homogeneity: $f_4 = \sum_{i,j=0}^{N-1} P_{ij}/{1+(i-j)^2}$ (6)

Correlation: $f_5 = \sum_{i,j=0}^{N-1} P_{ij} \frac{(i-\mu)(j-\mu)}{\sigma^2}$ (7)

Mean: $f_6 = \sum_{i,j=0}^{N-1} iP_{ij}$ (8)

Variance: $f_7 = \sum_{i,j=0}^{N-1} P_{i,j}(i-\mu)^2$ (9)

Where $P_{i,j}$ is the pixel value in position (i,j) in the output image and N is the number of gray levels in the output image. Now, for every entering image in the main box, features vector could be computed in every direction as follows: $F = (f_1, f_2, f_3, f_4, f_5, f_6, f_7)$.

Note that for every entering image into the main-box, there is 8 features vector, because for every images that entering into the second sub-box, GLCM's image can estimate in 8 directions. So for every image there are 56 computable features. $F_{final} = (f_{1,1}, f_{1,2}...f_{2,1}, f_{2,2}... f_{7,7})$.

In the results part, it has been shown that by using these features, stone textures classification could be done by high accuracy. From now on, the word "SEGL" would be use to name the approach offered in this article, which is the abbreviation of this words: "Statistical, Edge, GLCM and LBP".

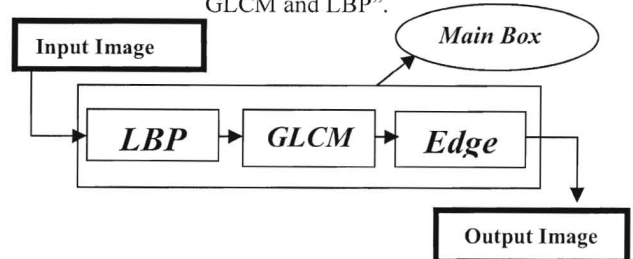

Figure6. Main box of proposed approach

## 6 Results

As it was mentioned in the introduction part, the main aim of this article is to offer a new approach for stone texture classification. Thus, to get results and observe the efficiency of proposed approach three kinds of stones named granite, travertine and hatchet, 60 images were taken by digital camera. It means 20 images of every model of stones. The images have been processed by the approach offered in this article, and the statistical features where computed for them according to section 5. To reduce the complexity and dataset's dimensions only GLCM was estimated in the $0^°$. Then there has been a dataset made consisting 60 instances, and 7 attributes. Every instance has a label which is the model of that stone. Finally, by using some of the classifiers such as KNN, NaiveBayes ,and LADTree, and by using N.Fold method, the accuracy of the dataset that has been made for stone textures is computed, which is shown in the table1. In second and third rows of the table1, the stone textures were classified just based on LBP and GLCM images to compare the proposed approach with some of the previous approaches in term of accuracy. As it is shown in table1, the accuracy of the new approach is much higher than other previous approaches. The main advantages of the proposed approach in this article can be mentioned in two points. 1-corresponding with near all of the classifiers, 2-Introducing a series of new features which are able to be computed for different applications.

| Classifier / Approach | 3NN | 5NN | Naive Bayes | LAD Tree |
|---|---|---|---|---|
| **SEGL** | **93.3 ± 0.8** | **93.6 ± 0.2** | **92 ± 0.2** | **90 ± 0.3** |
| *LBP* | 88.4 ± 0.8 | 82.6 ± 0.6 | 87.5± 0.5 | 81.4±0.8 |
| *GLCM* | 86.3 ± 0.6 | 85 ± 0.70 | 80.2± 0.4 | 79.8±0.7 |

Table1. Performance of Stone texture classification

## 7 Conclusion

The main purpose of this article was to offer an approach for stone texture classification. In this respect, the sections 2, 3 and 4 have algorithms of LBP, GLCM and edge estimations for the images and in section 5 the SEGL approach has been offered and described thoroughly. As it was described in SEGL method, first the input images were processed based on the main box, shown in Fig6, then, the statistical features were computed for the exiting images. In continuation, by using the exiting features in every image, we have made a dataset for three kinds of stones. Finally by using some classifiers and the N,Fold method we have classified all the items. So we see that all features which have computed by SEGL approach can use for classifying all kinds of stones by great accuracy.

## 8 Acknowledgment

The Author would likes to say thanks to Mr. Mohammad Abdollahi (Department of civil engineering) for his guidance to collate the database.